\documentclass[runningheads]{llncs}
\usepackage{graphicx}
\usepackage{color, soul}
	
\usepackage[table]{xcolor}
\usepackage{amsmath}

\usepackage{amsfonts}
\usepackage{multirow}
\usepackage{booktabs}
\usepackage{adjustbox}
\renewcommand{\arraystretch}{1.5}

\begin{document}
\title{ABODE-Net: An Attention-based Deep Learning Model for Non-intrusive Building Occupancy Detection Using Smart Meter Data}
\titlerunning{ABODE-Net: A Deep Learning Model for Building Occupancy Detection}
\author{Zhirui Luo\inst{1}\orcidID{0000-0002-3121-2597} 
\and
Ruobin Qi\inst{1}\orcidID{0000-0001-9072-9484} 
\and
Qingqing Li\inst{1}\orcidID{0000-0002-9006-3552} 
\and
Jun Zheng\inst{1}\orcidID{0000-0001-6727-5867} 
\and
Sihua Shao\inst{2}\orcidID{0000-0002-2831-9860}
}
\authorrunning{Z. Luo et al.}
%
\institute{Department of Computer Science and Engineering \\ 
New Mexico Institute of Mining and Technology, Socorro NM 87801, USA \\
\email{jun.zheng@nmt.edu}
\and
Department of Electrical Engineering \\ 
New Mexico Institute of Mining and Technology, Socorro NM 87801, USA
}
\maketitle              
\begin{abstract}
Occupancy information is useful for efficient energy management in the building sector. The massive high-resolution electrical power consumption data collected by smart meters in the advanced metering infrastructure (AMI) network make it possible to infer buildings' occupancy status in a non-intrusive way. In this paper, we propose a deep leaning model called ABODE-Net which employs a novel Parallel Attention (PA) block for building occupancy detection using smart meter data. The PA block combines the temporal, variable, and channel attention modules in a parallel way to signify important features for occupancy detection. We adopt two smart meter datasets widely used for building occupancy detection in our performance evaluation. A set of state-of-the-art shallow machine learning and deep learning models are included for performance comparison. The results show that ABODE-Net significantly outperforms other models in all experimental cases, which proves its validity as a solution for non-intrusive building occupancy detection.

\end{abstract}
\section{Introduction}
Recently, efficient energy management of buildings has attracted a lot of attention because of the significant potential for energy reduction. Building occupancy detection has many applications in this area such as improving the energy saving of building appliances and providing demand-response services for smart grids. The occupancy-based control of indoor Heating, Ventilation, and Air Conditioning (HVAC) systems and lighting in the buildings can lead up to 40\% reduction of the power consumption of the buildings and 76\% reduction of the power used for lighting, respectively \cite{Razavi2019}. Moreover, occupancy status information benefits demand-response services of smart grids by (i) determining users' peak demand periods \cite{Feng2020}, (ii) anticipating the willingness of deferring their consumption to off-peak hours \cite{albert2013}, and (iii) jointly optimizing the occupancy-based demand response and the thermal comfort for occupants in microgrids \cite{Korkas2016}.

Due to the wide deployment of Advanced Metering Infrastructures (AMI) globally, the massive high-resolution electrical power consumption data make non-intrusive building occupancy detection possible. With smart meters installed in customers' buildings, 
the building electricity usage data can be recorded and transmitted distantly in real-time for occupancy detection which does not require additional in-door sensors (e.g., environment measurement sensors or surveillance cameras). By using smart meter data, a number of studies were conducted that used data-driven machine learning models for building occupancy detection, such as support vector machine (SVM) \cite{chen2013nonintrusive,Kleiminger2015}, hidden Markov model (HMM) \cite{Kleiminger2015}, k-nearest neighbors (kNN) \cite{akbar2015,Kleiminger2015}, etc. Furthermore, a deep learning-based method proposed recently in \cite{Feng2020} sequentially stacks a convolutional neural network (CNN) and a bidirectional long short-term memory network (BiLSTM) to capture spatial and temporal patterns in the smart grid data, which outperforms other state-of-the-art methods. However, all of the aforementioned studies use a set of features manually extracted from the raw power consumption data. 
The goal of this paper is to develop an end-to-end deep learning model to automatically capture the discriminative information in the raw smart meter data to infer the building occupancy status.

Building occupancy detection based on raw smart meter data can be considered a time series classification (TSC) problem. In recent years, there were a considerable amount of studies that used deep learning to solve the challenging TSC problem \cite{ismail2019deep}. Specifically, CNN has demonstrated its powerful capability to solve multivariate TSC problems in many areas \cite{Chen2021,Karim2019,li2022anewdeep,luo2021}. Although recurrent neural networks (RNN), e.g. long short-term memory (LSTM), is good at capturing the time dependency, the lack of the parallel training ability caused by the recurrent calculation costs more computational power when scaling up. 
CNN can better utilize GPU parallelism since it does not have the recurrent structure like RNN. With the attention mechanism, the capability of a CNN-based model on capturing temporal and spatial patterns can be enhanced while keeping its advantage of parallel training.

In this paper, we propose an \textbf{\underline{A}}ttention-based  \textbf{\underline{B}}uilding \textbf{\underline{O}}ccupancy \textbf{\underline{De}}tection Deep Neural \textbf{\underline{Net}}work (ABODE-Net) which uses raw smart meter data as input. ABODE-Net utilizes a Fully Convolutional Network (FCN) block and a new Parallel-Attention (PA) block to learn both temporal and spacial patterns from smart meter data. The FCN uses three CNN layers to extract features automatically from power consumption readings and corresponding time information. The PA block combines temporal attention (TA), variable attention (VA), and Squeeze-and-Excitation (SE) modules to focus on important features by blending temporal, variable, and cross-channel information. The contributions of this paper are summarized as follows: (1) we propose an attention-based deep learning model called ABODE-Net for building occupancy detection in an end-to-end manner using raw smart meter data and corresponding time information; (2) we propose a novel lightweight PA block as a key component of ABODE-Net to capture discriminative information for building occupancy detection; and (3) we compare the performance of ABODE-Net with a set of state-of-the-art baseline methods using two popular smart meter datasets and prove that ABODE-Net is a viable solution for non-intrusive building occupancy detection using smart meter data.

This paper is organized as follows.
In Section \ref{sec:problem_statement}, we define the building occupancy detection problem. The proposed ABODE-Net for non-intrusive building occupancy detection using smart meter data is described in Section \ref{sec:method}. Section \ref{sec:experiment} presents the performance evaluation experiments and results. Finally, we conclude the paper in Section \ref{sec:conclusion}.

\section{Problem Definition}
\label{sec:problem_statement}
The building occupancy detection problem targeted in this paper is to infer the real-time occupancy status, $\mathbf{y}=\{y_i \mid  i \in[1,N]\}$, of a building from its historical power consumption data,
$\mathbf{X} = \{X_i \in \mathbb{R}^{1\times F \times T}  \mid i \in [1,N] \}$, where $N$, $F$ and $T$ are the number of samples, the number of features in each time step, and the number of time steps in the time window of a sample, respectively.
Specifically, we consider power consumption data and periodical time information as our features. The occupancy status $y_i$ of a building corresponding to the sample $X_i$ collected from the building is either vacant or occupied.
We seek to learn a set of trainable parameters, $\mathbf{W}$, of a deep neural network $M$ which predicts the building occupancy status $\hat{y}_i$ for the sample $X_i$. The prediction model for building occupancy detection can be written as:
\begin{equation}
    \hat{y}_i = M(\mathbf{W}, X_i)
\end{equation}

\section{Proposed Method}
\label{sec:method}
\begin{figure}[!htb]
    \centering
    \includegraphics[width=\textwidth]{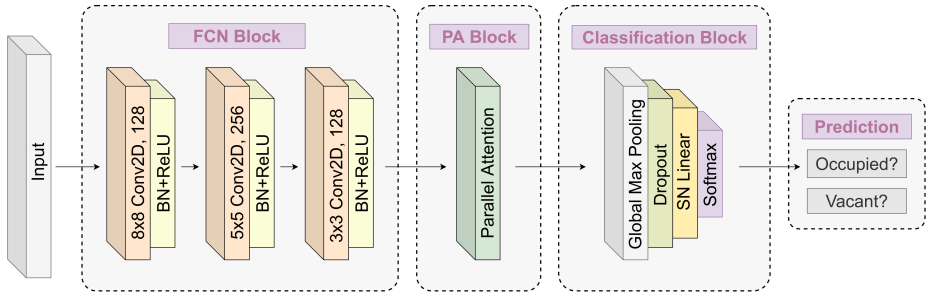}
    \caption{The architecture of ABODE-Net}
    \label{fig:overall_arch}
\end{figure}

The architecture of the proposed ABODE-Net is shown in Fig. \ref{fig:overall_arch}. ABODE-Net consists of three sequentially connected components: (i) an FCN block, (ii) a PA block, and (iii) a classification block. The FCN block serves as the function of automatic feature extraction. The PA block applies the attention mechanism to focus on good features and suppress poor features extracted from the FCN block. The classification block consists of a global pooling layer followed by a fully connected layer to generate the predicted possibility of occupancy for a given input. The symbols used in this section and their descriptions are listed in Table \ref{tab:allsymbols}.

\begin{table}[!htb]
\begin{center}
\caption{Symbols and their descriptions}
\label{tab:allsymbols}
\begin{adjustbox}{width=\textwidth}
\small
\begin{tabular}{ c l | c l }
\hline
\textbf{Symbol} & \textbf{Description} & \textbf{Symbol} & \textbf{Description} \\\hline
\textbf{W} & learnable weight parameters & Softmax & softmax activation \\
FC & fully-connected layer & $\delta$ & ReLU activation \\
Conv2d & convolution with 2D kernel & BN & batch normalization \\ 
Conv1x1 & Conv2d with $1\times 1$ kernel  & MLP & multilayer perceptron \\
GAP & global average pooling  & $\oplus$ & element-wise matrix addition \\
GMP & global max pooling & $\otimes$ & batch matrix multiplication \\
Sigmoid & Sigmoid activation & $\odot$ & element-wise multiplication \\ 
$\tanh$ & $\tanh$ activation & $\texttt{TRS}$ & tensor transpose operation\\
\hline
\end{tabular}
\end{adjustbox}
\end{center}
\end{table}

\subsection{FCN Block}
FCN has been proven to be an efficient feature extractor for many TSC tasks \cite{Hao2020,karim2017lstm,Karim2019}. In ABODE-Net, however, we use a different set of hyperparameters tailored for the building occupancy detection problem. For TSC tasks, FCN is usually constructed by using a large kernel size at the beginning to achieve enough perception fields, but with a shorter network depth compared to VGG-like networks \cite{simonyan2014very} and no pooling layer between convolution layers. In ABODE-Net, the FCN component consists of three sequentially connected basic blocks, B1 to B3. Each basic block sequentially stacks a Conv2d layer, a BN layer, and a ReLU activation function which can be written as:
\begin{align}
    h_l = B_l(x_l) = \sigma(\texttt{BN}(\texttt{Conv2d}(x_l))), l\in\{1,2,3\}
\end{align}
where $B_l$ indicates the $l$-th basic block, $x_l$ and $h_l$ are the input and output of $B_l$, respectively. Note that $x_1 \in \mathbb{R}^{1\times F\times T}$ is the input of the network which consists of the raw power consumption data and corresponding time information. The hyperparameters of Conv2d layers in the three basic blocks are listed in Table \ref{tab:fcn_hyper}.
\begin{table}[!htb]
    \centering
    \caption{Hyperparameters of Conv2d layers in three basic blocks of the FCN}
    \begin{adjustbox}{width=0.9\textwidth}
    \small
    \begin{tabular}{ccccc}
    \hline
        Basic Block & \# of filters & Kernel size & Padding & Stride \\\hline
        B1 & 128 & (8,8) & (3,3) & (4,4) \\ 
        B2 & 256 & (5,5) & (2,2) & (2,2) \\ 
        B3 & 128 & (3,3) & (1,1) & (1,1) \\ 
        \hline
    \end{tabular}
    \end{adjustbox}
    \label{tab:fcn_hyper}
\end{table}

\subsection{PA Block}
In the past decade, the attention mechanisms have played an increasingly important role in different deep learning applications such as computer vision \cite{fu2019dual,hu2018squeeze,woo2018cbam}, natural language processing (NLP) \cite{liu2019bidirectional,luong2015effective,sutskever2014sequence}, and TSC \cite{karim2017lstm,Karim2019,tang2016sequence}. An attention mechanism can dynamically weight important features and suppress trivial ones based on the input. The general form of the attention mechanism, $Attention(x)$, can be written as:
\begin{equation}
    Attention(x) = F(A(x),x)
\end{equation}
where $A(x)$ generates attention weights based on the input $x$ and $F(A(x),x)$ applies the attention weights $A(x)$ on the corresponding input $x$ to attend critical regions.

\begin{figure}[!htb]
    \centering
    \includegraphics[width=\textwidth]{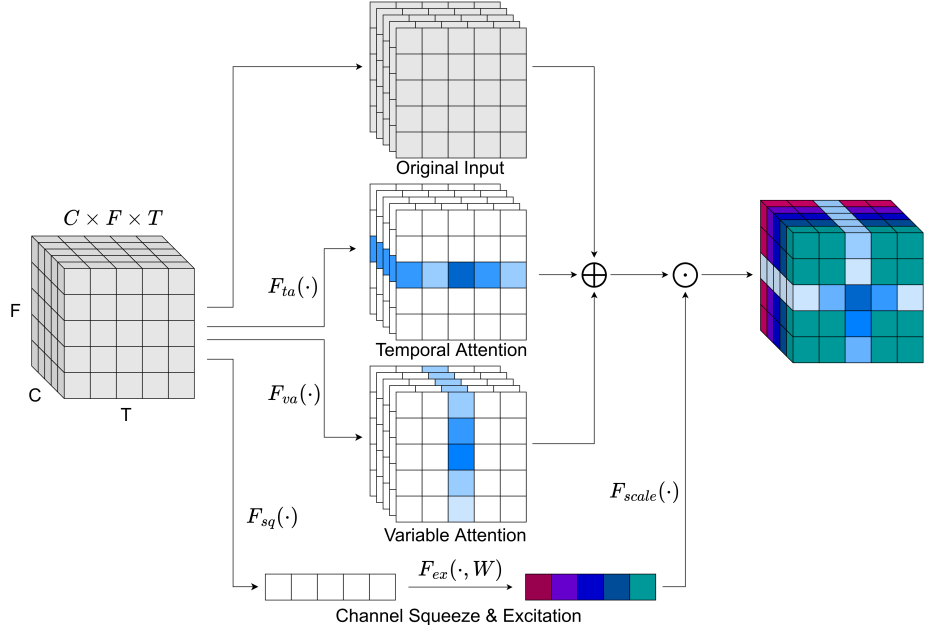}
    \caption{Illustration of the PA block}
    \label{fig:parallel_attention}
\end{figure}

A recent study classifies attention mechanisms into six categories: (i) channel attention, (ii) spatial attention, (iii) temporal attention, (iv) branch channel, (v) channel \& spatial attention, and (vi) spatial \& temporal attention \cite{guo2022attention}. On the other hand, a mechanism combining channel attention, variable attention, and temporal attention still remains unexplored in the TSC research area. We propose a novel PA block in ABODE-Net as shown in Fig. \ref{fig:parallel_attention} to combine the three kinds of attention modules in a parallel way to signify important features. 

The three attention modules in the PA block are the SE module, the VA module, and the TA module which are for cross-channel attention, variable attention, and temporal attention, respectively.  Given an intermediate feature map $H\in \mathbb{R}^{C\times F \times T}$ as input, PA parallelly infers a channel attention map $O_{\texttt{CA}} \in \mathbb{R}^{C\times F\times T}$, a variable attention map $O_{\texttt{VA}}\in \mathbb{R}^{C\times F\times T}$, and a temporal attention map $O_{\texttt{TA}} \in \mathbb{R}^{C\times F\times T}$. Once all attention maps are generated, they are combined according to Eq. (\ref{eq:attention_combination}) to obtain the final attended feature map $M_{\texttt{PA}} \in \mathbb{R}^{C\times F\times T}$.
\begin{align}
    M_{\texttt{PA}} = (H \oplus O_{\texttt{TA}} \oplus O_{\texttt{VA}} ) \odot O_{\texttt{CA}}
    \label{eq:attention_combination}
\end{align}

\subsubsection{Channel Attention Module}
To capture the cross-channel relationship, we adopt the SE block \cite{hu2018squeeze} as the channel attention module. SE exploits the channel dependencies by considering the channel-wise statistics from squeezed global variable and temporal information. The \textit{squeeze} operation $F_{sq}(\cdot)$ is defined by GAP. The \textit{excitation} operation $F_{ex}(\cdot, W)$ uses a simple gating mechanism with a sigmoid activation, where $W$ is the weights of a MLP. Finally, the output of the block is re-scaled back via the \textit{scale} operation $F_{scale}$. The SE module is formulated as follows:
\begin{align}
\label{SE_atten}
    H_{sq} &= F_{sq}(H) = GAP(H) \\
    H_{ex} &= F_{ex}(H_{sq},W) = \texttt{Sigmoid}(MLP(H_{sq}))\\
    &= \texttt{Sigmoid}(W_2\delta(W_1(GAP(H_{sq}))))\\
    O_{\texttt{CA}} &= F_{scale}(H_{ex})
\end{align}
where $W_1\in \mathbb{R}^{C\times \frac{C}{r}}$ and $W_2\in \mathbb{R}^{\frac{C}{r}\times C}$ are the trainable weights of two FC layers, $H_{sq}\in \mathbb{R}^{C}$ is the output of the \textit{squeeze} operation, $H_{ex}\in \mathbb{R}^{C}$ is the output of the \textit{excitation} operation, and $O_{\texttt{CA}} \in \mathbb{R}^{C\times F\times T}$ is the output of the \textit{scale} operation. Same as \cite{hu2018squeeze}, we set the reduction ratio $r$ to 16.

\subsubsection{Variable Attention Module}
The variable attention module in the PA block is implemented with the VA operation $F_{va}(\cdot)$. Given a local feature map $H\in \mathbb{R}^{C\times F \times T}$, the VA module first applies three convolutional layers with $1\times 1$ filters (Conv1x1) on $H$ to generate three feature maps $Q$, $K$, and $V$, respectively, where $\{Q,K\}\in \mathbb{R}^{C_1\times F\times T}$ and $\{V\}\in \mathbb{R}^{C_2\times F\times T}$. $C_1$ and $C_2$ are the number of channels, which are less than C for dimension reduction. We set $C_1$ to $C/8$ and $C_2$ to $C/2$.

We transpose $Q$ and $K$ before the attention weights are generated, where the transpose operation on tensor input swaps the axes of a given tensor. For the feature map $Q\in \mathbb{R}^{C_1\times F \times T}$, we transpose it into $Q'\in \mathbb{R}^{T\times F \times C_1}$. We transpose $K\in \mathbb{R}^{C_1\times F\times T}$ into $K'\in \mathbb{R}^{T \times C_1\times F}$. Following this, we apply a tanh activation function to $Q'$ and $K'$ separately, where we get $Q''$ and $K''$. For each time step $t \in T$, we multiply $Q''_{t}$ by $K''_{t}$ which results in a intermediate matrix $S_t\in \mathbb{R}^{F\times F}$. The softmax function is then applied to $S_t$ in order to generate the variable attention weight $A_{\texttt{VA}}\in \mathbb{R}^{T\times F\times F}$. 

We then calculate the attended value $D_{\texttt{VA}}=A_{\texttt{VA}}\otimes V'$, where $V'\in\mathbb{R}^{T\times F\times C_2}$ is transposed from $V\in \mathbb{R}^{C_2\times F\times T}$ and  $D_{\texttt{VA}}\in \mathbb{R}^{T\times F\times C_2}$. In order to reverse the attended feature map back to the same shape of $H$, we first transpose $D$ to $D'\in \mathbb{R}^{C_2\times F\times T}$. Next, we apply an additional Conv1x1 layer to increase the channels from $C_2$ back to $C$. Finally, we use a trainable scalar $\sigma_{\texttt{VA}}$ which adjusts the efficacy of the attended values to output $O_{\texttt{VA}}$. Therefore, $F_{va}(\cdot)$ can be formulated as: 
\begin{align}
    Q,K,V &= \texttt{Conv1x1}_{C_1}(H), \texttt{Conv1x1}_{C_1}(H), \texttt{Conv1x1}_{C_2}(H) \\
    Q' &= \texttt{TRS}_{C_1FT \rightarrow TFC_1}(Q)\\ 
    K' &= \texttt{TRS}_{C_1FT\rightarrow TC_1F}(K) \\
    V' &= \texttt{TRS}_{C_2FT\rightarrow TFC_2}(V)\\
    A_{\texttt{VA}} &= \texttt{Softmax}(\tanh(Q')\otimes \tanh(K')) \\
    D_{\texttt{VA}} &= A_{\texttt{VA}}\otimes V'\\
    O_{\texttt{VA}} &= \sigma_{\texttt{VA}} \cdot \texttt{Conv1x1}_{C}(\texttt{TRS}_{TFC_2\rightarrow C_2FT}(D_{\texttt{VA}}))
\end{align}
where $\texttt{TRS}$ is the transpose operation and its subscript indicates the axes of the given tensor swapped.
$\texttt{Conv1x1}_{c}$ is the convolutional layer with the $1\times1$ filter and $c$ output channels.

\subsubsection{Temporal Attention Module}
The temporal dependencies are captured via the TA operation $F_{ta}(\cdot)$. The TA module is similar to the VA module with some differences in transposing $Q,K,V$ feature maps. The resulting attention weight of TA is $A_{\texttt{TA}} \in \mathbb{R}^{F\times T\times T}$ while the attention weight for VA is $A_{\texttt{VA}}\in \mathbb{R}^{T\times F \times F}$. $F_{ta}(\cdot)$ is formulated as follows: 
\begin{align}
    Q,K,V &= \texttt{Conv1x1}_{C_1}(H), \texttt{Conv1x1}_{C_1}(H), \texttt{Conv1x1}_{C_2}(H) \\
    Q' &= \texttt{TRS}_{C_1FT \rightarrow FTC_1}(Q)\\ 
    K' &= \texttt{TRS}_{C_1FT\rightarrow FC_1T}(K) \\
    V' &= \texttt{TRS}_{C_2FT\rightarrow FTC_2}(V)\\
    A_{\texttt{TA}} &= \texttt{Softmax}(\tanh(Q')\otimes \tanh(K')) \\
    D_{\texttt{TA}} &= A_{\texttt{TA}}\otimes V'\\
    O_{\texttt{TA}} &= \sigma_{\texttt{TA}} \cdot \texttt{Conv1x1}_{C}(\texttt{TRS}_{FTC_2\rightarrow C_2FT}(D_{\texttt{TA}}))
\end{align}

\subsection{Classification Block}
Given the attended feature map $M_\texttt{PA}$, we use a GMP layer followed by an FC layer and a softmax function to predict the occupancy status. The classification block can be formulated as:
\begin{align}
     h_{fc} &= \texttt{SN}(W_{c}(\texttt{GMP}(M_\texttt{PA}))+b_c)\\
     \hat{\texttt{y}} &= \texttt{Softmax}(h_{fc})
\end{align}
where $W_c\in \mathbb{R}^{128\times N_c}$ and $b_c$ are the weight and bias of the FC layer, respectively. The output of the FC layer is $h_{fc} \in \mathbb{R}^{N_c}$ and $\hat{\texttt{y}} \in \mathbb{R}^{N_c}$ is the vector of predicted probabilities, where $N_c$ is the number of classes.
The spectral normalization (SN) is utilized to stabilize the FC layer, which is a normalization technique proposed in \cite{miyato2018spectral} to stabilize the training of the discriminator in generative adversarial networks (GANs). Unlike input based regularizations, e.g. BN, SN does not depend on the space of data distribution, instead it normalizes the weight matrices.

\subsection{Model Training}
\label{sec:model_training}
The three trainable components of ABODE-Net are trained together. Since the building occupancy detection task is a classification problem, we use the negative log likelihood loss ($\texttt{NLL}_{loss}$) shown in Eq. (\ref{eq:loss}) for training:
\begin{align}
    \texttt{NLL}_{loss}(y, \hat{y}) = -\sum_{i=1}^{N_c} y_i \ln \hat{y}_i
    \label{eq:loss}
\end{align}
where $N_c$ is the number of classes, $y$ is the one-hot encoding of the ground truth label, and $y_i$ and $\hat{y}_i$ are the actual and predicted probabilities of the label being $i$th class, respectively. The predicted occupancy status is the class whose predicted probability is the highest among all classes.

We use Adam with a learning rate of 1e-3 as the optimizer to learn all trainable parameters of ABODE-Net. To prevent over-fitting, weight decay is set to 5e-4. The max training epoch is 100. We use a scheduler with a learning rate that decreases following the value of cosine function between the initial learning rate and 0 with a warm-up period of 7 epochs.

\section{Performance Evaluation and Results}
\label{sec:experiment}
\subsection{Smart Meter Datasets}
To evaluate the performance of ABODE-Net, two smart meter datasets widely used for non-intrusive building occupancy detection, the Electricity Consumption and Occupancy (ECO) dataset \cite{beckel2014eco} and the Non-Intrusive Occupancy Monitoring (NIOM) dataset \cite{chen2013nonintrusive}, are adopted in our study. 
The ECO dataset was collected from five houses in summer and winter. The NIOM dataset was collected from two houses, Home-A in both spring and summer and Home-B only in summer. We use the power consumption readings, i.e. the \textit{powerallphases} of the ECO dataset and the power usage trace of the NIOM dataset, and their corresponding occupancy statuses for our experiments. The occupancy status of the NIOM dataset is for two occupants. However, we consider a home to be occupied when at least one occupant is at home. 
The sampling rates of the two datasets are different. The ECO dataset has a sampling rate of one sample per second, while the NIOM dataset has a lower sampling rate of one sample per minute. Thus, we aggregate the samples of every 60s of the ECO dataset into one sample via averaging.

In addition to power consumption information, time information is also important for occupancy detection because the life style of a household usually depends on the time of a day and the day of a week. We add time of the day $P_{time}$ and day of the week $P_{day}$ to model the temporal dependency and life-style cycle. $P_{time}$ is the timestamp ranging from 0 to 1439 corresponding to each minute between 00:00 to 23:59. $P_{day}$ is the day of a week ranging from 0 to 6 corresponding to Monday to Sunday. For both datasets, the input features of a sample are generated from a 60-min power consumption readings and their corresponding time information, $P_{time}$ and $P_{day}$, resulting in a $3 \times 60$ feature matrix. We add a dummy dimension to the feature matrix to form the input of the FCN of shape $1\times 3 \times 60$. The label of an input is the majority class (occupied or vacant) of samples in the 60-min segment.

\subsection{Data Preparation}
For a fair comparison, the datasets are quality-controlled with two criteria \cite{Feng2020}: (1) the data length should be more than 900 samples, and (2) the samples of each class should be more than 10\% of all samples. Based on the criteria, four periods of the three houses in the ECO dataset and all periods of the two houses in the NIOM dataset are qualified which are shown in Table \ref{tab:datasets}.  Each qualified period of a house is used as a case for performance evaluation.
\begin{table}[!ht]
    \centering
    \caption{Qualified cases of the two datasets}
    \label{tab:datasets}
    \begin{adjustbox}{width=0.9\textwidth}
    \small
    \begin{tabular}{cccc}
        \hline
         Case & Household-Period  & Occupied/Vacant & Total \# of samples \\\hline
         ECO-1 & 01-Summer & 769/168 & 937 \\ 
         ECO-2 & 01-Winter & 834/270 & 1104 \\
         ECO-3 & 02-Winter & 769/311 & 1080 \\
         ECO-4 & 03-Summer & 1038/330 & 1368\\
         NIOM-1 & HomeA-Spring & 125/43 & 168 \\ 
         NIOM-2 & HomeA-Summer & 142/26 & 168 \\
         NIOM-3 & HomeB-Summer & 125/43 & 168 \\
         \hline
    \end{tabular}
    \end{adjustbox}
\end{table}

As shown in Table \ref{tab:datasets}, all cases have significantly more occupied samples than vacant samples which result in imbalanced datasets for performance evaluation.
Therefore, we use the random oversampling method implemented by \texttt{Imbalanced-learn} \cite{guillaume2017imb} to over-sample the minority vacant class. The random oversampling method randomly select samples in the minority class with replacement. 

In an evaluation experiment, the data of each case are randomly divided into training, validation, and test sets by a ratio of 3:1:1. The experiment is repeated 10 times randomly by using different random seeds. We apply the min-max normalization to scale each feature of the data into the range between zero to one. 
After the normalization, we oversample the minority class of the training set with the validation and test sets untouched.

\subsection{Baseline Models}
We compare ABODE-Net with a collection of state-of-the-art shallow ML and DL models which have been applied for building occupancy detection. All models are implemented using Python 3.8. The DL models and ABODE-Net are implemented with the PyTorch framework.
The included shallow ML models are: kNN \cite{akbar2015,Kleiminger2015}, GMM \cite{Kleiminger2015}, and SVM \cite{chen2013nonintrusive,Kleiminger2015}, which are implemented with the \texttt{scikit-learn} library. We use the grid search to find the best hyperparameters of those models. 

The DL models included in our experiments are CNN, LSTM, and CNN-BiLSTM of \cite{Feng2020}. We re-implement CNN-BiLSTM which uses the same features as ABODE-Net.
The CNN model has the same architecture as the CNN block of CNN-BiLSTM. The LSTM model consists of two LSTM layers with 50 hidden neurons and bias weights, and its classification block contains a dropout layer with 40\% dropping rate followed by spectral-normalized FC layer. We train baseline DL models the same way as we train ABODE-Net which is discussed in Section \ref{sec:model_training}. Specifically, the hyperparameters of a deep model achieving the highest F1 score on the validation phase are used for testing.

\subsection{Performance Metrics}
Two popular metrics, accuracy and F1 score, are used for evaluating the performance of all models. By denoting the two occupancy statuses, occupied and vacant, as positive and negative, respectively, the two performance metrics are defined as:
\begin{align}
    & Accuracy = \frac{T_p + T_n}{T_p+T_n+F_p+F_n}\\
    & Precision = \frac{T_p}{T_p+F_p}\\
    & Recall = \frac{T_p}{T_p+T_n}\\
    & F_1 = 2\cdot \frac{precision\cdot recall}{precision + recall}
\end{align}
where $T_p$, $T_n$, $F_p$, and $F_n$ are true positives, true negatives, false positives, and false negatives, respectively.

\subsection{Results}
The performance evaluation results in term of averaging accuracy and F1 score over 10 trials are shown in Tables \ref{tab:result_eco_random_cases} and \ref{tab:result_niom_random_cases} for the cases of the ECO and NIOM datasets, respectively. As shown in Table \ref{tab:result_eco_random_cases}, ABODE-Net outperforms all baseline models for every case of the ECO dataset, whose average accuracy and F1 score over all cases are $0.8649$ and $0.8198$, respectively. 
It can be seen that the three shallow ML models have significantly worse performance than DL models. This shows that DL models have better capability of capturing time dependencies and spatial patterns in time series data. Another reason is that these shallow ML models are affected by the curse of dimensionality due to the high dimensional data. 
Similar results can be observed in Table \ref{tab:result_niom_random_cases} for the NIOM dataset. ABODE-Net achieves the best average accuracy and F1 score over all cases among all models. It only has a slightly lower F1 score than LSTM for case 3. On the other hand, LSTM has much worse performance than ABODE-Net in other two cases. 
Overall, the performance evaluation results demonstrate that ABODE-Net is more capable than the state-of-the-art baseline models in terms of utilizing the temporal and spatial information in smart meter data to detect building occupancy.

\begin{table}[!ht]
\centering
\arrayrulewidth=0.5pt
\setlength\tabcolsep{1.5pt}
\renewcommand{\arraystretch}{1.1}
\caption{Performance evaluation results for the ECO dataset}
\label{tab:result_eco_random_cases}
\begin{adjustbox}{width=\textwidth}
\small
\begin{tabular}{|cc|ccccccc|}
\hline
\multirow[c]{2}{*}{\textbf{Case}} & \multirow[c]{2}{*}{\textbf{Metric}} & \multicolumn{7}{c|}{\textbf{Model}} \\
{} & {} & {ABODE-Net} & {kNN} & {GMM} & {SVM} & {CNN} & {LSTM} & {CNN\_BiLSTM} \\
\hline
\multirow[c]{2}{*}{1} & ACC & {\cellcolor[HTML]{97D385}} \color[HTML]{000000} \bfseries 0.8585 & {\cellcolor[HTML]{A7DB8C}} \color[HTML]{000000} 0.7766 & {\cellcolor[HTML]{B3E091}} \color[HTML]{000000} 0.7133 & {\cellcolor[HTML]{A6DA8B}} \color[HTML]{000000} 0.7851 & {\cellcolor[HTML]{9AD587}} \color[HTML]{000000} 0.8426 & {\cellcolor[HTML]{9FD788}} \color[HTML]{000000} 0.8154 & {\cellcolor[HTML]{9CD687}} \color[HTML]{000000} 0.8324 \\
 & F1 & {\cellcolor[HTML]{A6DA8B}} \color[HTML]{000000} \bfseries 0.7832 & {\cellcolor[HTML]{B3E091}} \color[HTML]{000000} 0.7153 & {\cellcolor[HTML]{C0E597}} \color[HTML]{000000} 0.6483 & {\cellcolor[HTML]{B2DF90}} \color[HTML]{000000} 0.7217 & {\cellcolor[HTML]{ACDD8E}} \color[HTML]{000000} 0.7562 & {\cellcolor[HTML]{ACDD8E}} \color[HTML]{000000} 0.7509 & {\cellcolor[HTML]{ABDC8D}} \color[HTML]{000000} 0.7621 \\
\cline{1-9}
\multirow[c]{2}{*}{2} & ACC & {\cellcolor[HTML]{9DD688}} \color[HTML]{000000} \bfseries 0.8208 & {\cellcolor[HTML]{C9E99C}} \color[HTML]{000000} 0.5900 & {\cellcolor[HTML]{B2DF90}} \color[HTML]{000000} 0.7262 & {\cellcolor[HTML]{BCE395}} \color[HTML]{000000} 0.6701 & {\cellcolor[HTML]{9FD788}} \color[HTML]{000000} 0.8131 & {\cellcolor[HTML]{9FD788}} \color[HTML]{000000} 0.8190 & {\cellcolor[HTML]{ABDC8D}} \color[HTML]{000000} 0.7584 \\
 & F1 & {\cellcolor[HTML]{A9DB8C}} \color[HTML]{000000} \bfseries 0.7719 & {\cellcolor[HTML]{CFEC9E}} \color[HTML]{000000} 0.5553 & {\cellcolor[HTML]{B6E192}} \color[HTML]{000000} 0.7024 & {\cellcolor[HTML]{C0E597}} \color[HTML]{000000} 0.6414 & {\cellcolor[HTML]{ABDC8D}} \color[HTML]{000000} 0.7654 & {\cellcolor[HTML]{ABDC8D}} \color[HTML]{000000} 0.7650 & {\cellcolor[HTML]{D7EFA2}} \color[HTML]{000000} 0.5147 \\
\cline{1-9}
\multirow[c]{2}{*}{3} & ACC & {\cellcolor[HTML]{89CE80}} \color[HTML]{000000} \bfseries 0.9218 & {\cellcolor[HTML]{9CD687}} \color[HTML]{000000} 0.8292 & {\cellcolor[HTML]{ABDC8D}} \color[HTML]{000000} 0.7588 & {\cellcolor[HTML]{97D385}} \color[HTML]{000000} 0.8537 & {\cellcolor[HTML]{89CE80}} \color[HTML]{000000} 0.9194 & {\cellcolor[HTML]{89CE80}} \color[HTML]{000000} 0.9185 & {\cellcolor[HTML]{8BCE81}} \color[HTML]{000000} 0.9111 \\
 & F1 & {\cellcolor[HTML]{8DCF81}} \color[HTML]{000000} \bfseries 0.9050 & {\cellcolor[HTML]{A1D889}} \color[HTML]{000000} 0.8081 & {\cellcolor[HTML]{C3E698}} \color[HTML]{000000} 0.6287 & {\cellcolor[HTML]{9CD687}} \color[HTML]{000000} 0.8349 & {\cellcolor[HTML]{8DCF81}} \color[HTML]{000000} 0.9004 & {\cellcolor[HTML]{8DCF81}} \color[HTML]{000000} 0.9015 & {\cellcolor[HTML]{93D284}} \color[HTML]{000000} 0.8672 \\
\cline{1-9}
\multirow[c]{2}{*}{4} & ACC & {\cellcolor[HTML]{97D385}} \color[HTML]{000000} \bfseries 0.8584 & {\cellcolor[HTML]{B9E294}} \color[HTML]{000000} 0.6818 & {\cellcolor[HTML]{BEE596}} \color[HTML]{000000} 0.6562 & {\cellcolor[HTML]{BDE496}} \color[HTML]{000000} 0.6569 & {\cellcolor[HTML]{9DD688}} \color[HTML]{000000} 0.8248 & {\cellcolor[HTML]{AEDD8E}} \color[HTML]{000000} 0.7482 & {\cellcolor[HTML]{A2D88A}} \color[HTML]{000000} 0.8036 \\
 & F1 & {\cellcolor[HTML]{9FD788}} \color[HTML]{000000} \bfseries 0.8191 & {\cellcolor[HTML]{BEE596}} \color[HTML]{000000} 0.6497 & {\cellcolor[HTML]{C1E698}} \color[HTML]{000000} 0.6367 & {\cellcolor[HTML]{C1E698}} \color[HTML]{000000} 0.6362 & {\cellcolor[HTML]{A4D98A}} \color[HTML]{000000} 0.7923 & {\cellcolor[HTML]{B3E091}} \color[HTML]{000000} 0.7181 & {\cellcolor[HTML]{ABDC8D}} \color[HTML]{000000} 0.7655 \\
\cline{1-9}
\multirow[c]{2}{*}{Average} & ACC & {\cellcolor[HTML]{95D385}} \color[HTML]{000000} \bfseries 0.8649 & {\cellcolor[HTML]{B2DF90}} \color[HTML]{000000} 0.7194 & {\cellcolor[HTML]{B3E091}} \color[HTML]{000000} 0.7136 & {\cellcolor[HTML]{AFDE8F}} \color[HTML]{000000} 0.7415 & {\cellcolor[HTML]{98D486}} \color[HTML]{000000} 0.8500 & {\cellcolor[HTML]{9DD688}} \color[HTML]{000000} 0.8253 & {\cellcolor[HTML]{9DD688}} \color[HTML]{000000} 0.8264 \\
 & F1 & {\cellcolor[HTML]{9FD788}} \color[HTML]{000000} \bfseries 0.8198 & {\cellcolor[HTML]{B9E294}} \color[HTML]{000000} 0.6821 & {\cellcolor[HTML]{BEE596}} \color[HTML]{000000} 0.6540 & {\cellcolor[HTML]{B5E092}} \color[HTML]{000000} 0.7085 & {\cellcolor[HTML]{A2D88A}} \color[HTML]{000000} 0.8036 & {\cellcolor[HTML]{A6DA8B}} \color[HTML]{000000} 0.7839 & {\cellcolor[HTML]{B1DF90}} \color[HTML]{000000} 0.7274 \\
\hline
\end{tabular}
\end{adjustbox}
\end{table}

\begin{table}[!ht]
    \centering
    \arrayrulewidth=0.5pt
\setlength\tabcolsep{1.5pt}
\renewcommand{\arraystretch}{1.1}
    \caption{Performance evaluation results for the NIOM dataset}
    \label{tab:result_niom_random_cases}
\begin{adjustbox}{width=\textwidth}
\small
\begin{tabular}{|cc|ccccccc|}
\hline
\multirow[c]{2}{*}{\textbf{Case}} & \multirow[c]{2}{*}{\textbf{Metric}} & \multicolumn{7}{c|}{\textbf{Model}} \\
{} & {} & {ABODE-Net} & {kNN} & {GMM} & {SVM} & {CNN} & {LSTM} & {CNN\_BiLSTM} \\
\hline
\multirow[c]{2}{*}{1} & ACC & {\cellcolor[HTML]{89CE80}} \color[HTML]{000000} \bfseries 0.9206 & {\cellcolor[HTML]{98D486}} \color[HTML]{000000} 0.8500 & {\cellcolor[HTML]{A2D88A}} \color[HTML]{000000} 0.8029 & {\cellcolor[HTML]{97D385}} \color[HTML]{000000} 0.8529 & {\cellcolor[HTML]{8ED082}} \color[HTML]{000000} 0.8912 & {\cellcolor[HTML]{ACDD8E}} \color[HTML]{000000} 0.7529 & {\cellcolor[HTML]{8DCF81}} \color[HTML]{000000} 0.9000 \\
 & F1 & {\cellcolor[HTML]{8DCF81}} \color[HTML]{000000} \bfseries 0.9007 & {\cellcolor[HTML]{9DD688}} \color[HTML]{000000} 0.8248 & {\cellcolor[HTML]{C3E698}} \color[HTML]{000000} 0.6300 & {\cellcolor[HTML]{9DD688}} \color[HTML]{000000} 0.8207 & {\cellcolor[HTML]{95D385}} \color[HTML]{000000} 0.8601 & {\cellcolor[HTML]{B2DF90}} \color[HTML]{000000} 0.7202 & {\cellcolor[HTML]{95D385}} \color[HTML]{000000} 0.8633 \\
\cline{1-9}
\multirow[c]{2}{*}{2} & ACC & {\cellcolor[HTML]{8DCF81}} \color[HTML]{000000} \bfseries 0.9000 & {\cellcolor[HTML]{9DD688}} \color[HTML]{000000} 0.8235 & {\cellcolor[HTML]{92D183}} \color[HTML]{000000} 0.8824 & {\cellcolor[HTML]{9CD687}} \color[HTML]{000000} 0.8353 & {\cellcolor[HTML]{93D284}} \color[HTML]{000000} 0.8706 & {\cellcolor[HTML]{AFDE8F}} \color[HTML]{000000} 0.7412 & {\cellcolor[HTML]{9CD687}} \color[HTML]{000000} 0.8294 \\
 & F1 & {\cellcolor[HTML]{A9DB8C}} \color[HTML]{000000} \bfseries 0.7730 & {\cellcolor[HTML]{AFDE8F}} \color[HTML]{000000} 0.7388 & {\cellcolor[HTML]{C3E698}} \color[HTML]{000000} 0.6267 & {\cellcolor[HTML]{B1DF90}} \color[HTML]{000000} 0.7276 & {\cellcolor[HTML]{ACDD8E}} \color[HTML]{000000} 0.7513 & {\cellcolor[HTML]{C4E799}} \color[HTML]{000000} 0.6198 & {\cellcolor[HTML]{B1DF90}} \color[HTML]{000000} 0.7269 \\
\cline{1-9}
\multirow[c]{2}{*}{3} & ACC & {\cellcolor[HTML]{86CC7F}} \color[HTML]{000000} \bfseries 0.9324 & {\cellcolor[HTML]{8DCF81}} \color[HTML]{000000} 0.9029 & {\cellcolor[HTML]{9CD687}} \color[HTML]{000000} 0.8353 & {\cellcolor[HTML]{88CD7F}} \color[HTML]{000000} 0.9294 & {\cellcolor[HTML]{89CE80}} \color[HTML]{000000} 0.9206 & {\cellcolor[HTML]{86CC7F}} \color[HTML]{000000} \bfseries 0.9324 & {\cellcolor[HTML]{88CD7F}} \color[HTML]{000000} 0.9265 \\
 & F1 & {\cellcolor[HTML]{89CE80}} \color[HTML]{000000} 0.9190 & {\cellcolor[HTML]{90D083}} \color[HTML]{000000} 0.8878 & {\cellcolor[HTML]{B5E092}} \color[HTML]{000000} 0.7041 & {\cellcolor[HTML]{89CE80}} \color[HTML]{000000} 0.9162 & {\cellcolor[HTML]{8DCF81}} \color[HTML]{000000} 0.9047 & {\cellcolor[HTML]{89CE80}} \color[HTML]{000000} \bfseries 0.9194 & {\cellcolor[HTML]{8BCE81}} \color[HTML]{000000} 0.9091 \\
\cline{1-9}
\multirow[c]{2}{*}{Average} & ACC & {\cellcolor[HTML]{89CE80}} \color[HTML]{000000} \bfseries 0.9176 & {\cellcolor[HTML]{97D385}} \color[HTML]{000000} 0.8588 & {\cellcolor[HTML]{9AD587}} \color[HTML]{000000} 0.8402 & {\cellcolor[HTML]{93D284}} \color[HTML]{000000} 0.8725 & {\cellcolor[HTML]{8ED082}} \color[HTML]{000000} 0.8941 & {\cellcolor[HTML]{A1D889}} \color[HTML]{000000} 0.8088 & {\cellcolor[HTML]{90D083}} \color[HTML]{000000} 0.8853 \\
 & F1 & {\cellcolor[HTML]{95D385}} \color[HTML]{000000} \bfseries 0.8643 & {\cellcolor[HTML]{9FD788}} \color[HTML]{000000} 0.8171 & {\cellcolor[HTML]{BEE596}} \color[HTML]{000000} 0.6536 & {\cellcolor[HTML]{9DD688}} \color[HTML]{000000} 0.8215 & {\cellcolor[HTML]{9AD587}} \color[HTML]{000000} 0.8387 & {\cellcolor[HTML]{ACDD8E}} \color[HTML]{000000} 0.7531 & {\cellcolor[HTML]{9CD687}} \color[HTML]{000000} 0.8331 \\
\hline
\end{tabular}
\end{adjustbox}
\end{table}

\section{Conclusion}
\label{sec:conclusion}
In this paper, we propose an attention-based deep learning model called ABODE-Net to infer the building occupancy status in an end-to-end manner by using the raw smart meter data and corresponding time information. ABODE-Net consists of a three-layer FCN block, a novel light-weight PA block, and a spectral-normalized classification block. 
The proposed PA block captures discriminative information for occupancy detection through a combination of temporal, variable, and channel attentions. Two smart meter datasets widely used for building occupancy detection are adopted for performance evaluation. The experimental results demonstrate that ABODE-Net achieves significantly better performance than the state-of-the-art baseline methods. 

\subsubsection*{Acknowledgements.}
This work was supported in part by the National Science Foundation under EPSCoR Cooperative Agreement OIA-1757207 and in part by the Institute for Complex Additive Systems Analysis (ICASA) of New Mexico Institute of Mining and Technology.

%
%
%
 \bibliographystyle{splncs04}
 \bibliography{refs}

\end{document}